# Using Unlabeled Data for Increasing Low-Shot Classification Accuracy of Relevant and Open-Set Irrelevant Images

Spiridon Kasapis [α], Geng Zhang [σ], Nickolas Vlahopoulos [ρ] & Jonathon M. Smereka [ω]

*Abstract-* In search, exploration, and reconnaissance tasks performed with autonomous ground vehicles, an image classification capability is needed for specifically identifying targeted objects (relevant classes) and at the same time recognize when a candidate image does not belong to anyone of the relevant classes (irrelevant images). In this paper, we present an open-set low-shot classifier that uses, during its training, a modest number (less than 40) of labeled images for each relevant class, and unlabeled irrelevant images that are randomly selected at each epoch of the training process. The new classifier is capable of identifying images from the relevant classes, determining when a candidate image is irrelevant, and it can further recognize categories of irrelevant images that were not included in the training (unseen). The proposed low-shot classifier can be attached as a top layer to any pre-trained feature extractor when constructing a Convolutional Neural Network.

## I. Introduction

Extensive research in the field of machine learning has been progressively improving the performance of object recognition algorithms which achieve impressive results on a variety of multi-class classification tasks [15, 17, 24]. Especially in search, exploration, and reconnaissance applications where object recognition methods have been concentrated on a closed-set setting where all testing samples belong to one of the classes that the classifier has been trained on [29]. The limited finite number of classes which are the target of inspection need to be detected out of the infinite object classes that are encountered in unconstrained environment.

To tackle this challenge, efforts have been made to endow Convolutional Neural Networks (CNNs) the innate human brain capability to identify objects they are trained on while deliberately discarding objects of no interest. Lately, the introduction of open-set classification [20, 31, 30] has introduced an ability to correctly identify images as unknown test objects that do not belong to any known classes, as opposed to falsely classifying them in one of the known classes (i.e., classes that the model has been trained on). More specifically, [28, 10] defines open-set classification as the problem of balancing the known space (specialization) and unknown open space (generalization) of the model. Examples such as out-of distribution detection [18] and realistic classification [26] show the interest in the concept of open-set recognition [4] while showing that CNNs can be trained to reject examples that have not been seen during training or are too hard to classify.

Recently, works on video object discovery [33] go against the closed-set assumption that each image during inference belongs to one of the fixed number of relevant classes. In [33] the terminology of relevant and irrelevant is introduced and is used in this paper since it aligns with the definitions stated in the Abstract. In most real-life applications this closedset assumption is uncommon and ideal, therefore recently proposed methods [4] are subject to an open-set condition where images not seen during training should be classified into irrelevant or unseen classes. Consequentially, in this work we introduce the splitting of testing samples in three categories: (a) relevant; labeled samples used during train-

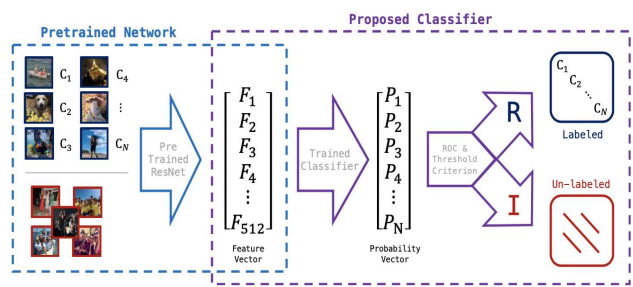

*Figure 1:* Schematic of the two parts of our network. We feed to the Pretrained Network labeled "Relevant" images and unlabeled "Irrelevant" images. For each image our proposed classifier produces class score vectors that get classified using a threshold criterion and Receiver Operating Characteristic (ROC), with accuracy much greater than already existing techniques, especially for the irrelevant dataset.

*Author α ρ: University of Michigan. e-mails: skasapis@umich.edu, nickvl@umich.edu*
*Author σ: Michigan Engineering Services, e-mail: gengz@miengsrv.com*
*Author Ω: US Army DEVCOM GVSC.*
*e-mail: jonathon.m.smereka.civ@army.mil*

ing, (b) irrelevant; unlabeled samples used during training and (c) irrelevant but also unseen; for categories of images that are not seen during training and should be identified as irrelevant.

Another challenge the visual recognition community faces is the absence of labeled examples. Especially in military applications having large labeled datasets is an unreal expectation as needs and mission tools used for search and reconnaissance evolve. An open-set recognizer will face limitations such as the absence of large amounts of training samples, thus an open-set recognition technique that simultaneously supports the few-shot setting is needed. Therefore, in this paper we propose a low-shot solution to the problem of open-set recognition which considers exclusively the classification layer of a CNN.

Specifically, we present an approach on significantly improving the performance of a simple, time efficient, one-layer classifier on recognising labeled (relevant) images along with non-labeled (irrelevant and unseen as mentioned above) images (Figure 1). The ability to specifically recognize a number (of the order of 50) of relevant classes and also identify when an image does not belong to any of them in a labelinexpensive way is one of the main motivations for low-shot open-set recognition.

Efforts with similar goals have been concentrated on the training of the entire CNN. For example, the PEELER algorithm [25] combines the random selection of a set of novel classes per episode, a loss that maximizes the posterior entropy for examples of those classes, and a new metric learning formulation in order to train the weights of a CNN in such a manner that it can recognize images of a limited amount of classes ($\leq$ 20) that are unseen during training. Dhamija et al. [7] proposes the introduction of two loss functions that are designed to maximize entropy for unknown inputs while increasing separation in deep feature space by modifying magnitudes of known and unknown samples. Although the work of Dhamija et al. introduces the concept of unknown sample recognition like we do, the number of recognizable classes is still very limited compared to the testing done in our method.

Both of the aforementioned algorithms train the entirety of the CNN, unlike the methods proposed by Kozerawski and Turk [23] which can augment any few shot learning method without requiring retraining in order to work in a few-shot multiclass open-set setting. Although not concerned with one-class classification, a similar approach is followed in our work too, where we utilize a pre-trained feature extractor (such as the ones publicly available by PyTorch[1]) and propose an independent open-set low-shot classification method which can augment any existing feature extractor.

To explore the open-set low-shot problem in a holistic, non-specific and easily applicable way, we concentrate only on the training of the classification matrix (matrix used to turn the feature vector to a probability vector in Figure 1), using the pre-trained ResNet feature extractors [15] discussed in Section 2. We reduce the image matrices to feature vectors [8, 1] which are then used in Section 4 to train the classifier with the help of the analytic derivative of our loss function and a unique, partially labeled, target matrix. In Section 5, we use the classification variability statistics and a Receiver Operating Characteristic (ROC) curve as a method to calculate threshold scores for each relevant class. An approach that uses random selections of unlabeled irrelevant images during each epoch of the classifier training is introduced. Testing datasets are used in Section 3 for determining the ability to effectively classify all Relevant, Irrelevant and Unseen datasets. In the last Section we make some closing remarks on our work presented in this paper.

In summary, the contributions of this work are the following:

- We present a novel open-set low-shot (OSLS) classification method which can be added as the top layer to any pre-trained feature extractor in order to create a CNN that can classify images in relevant classes and also determine if an image does not belong to any of the relevant classes.
- The OSLS Classifier yields improved classification performance compared to classifiers that either do not use unlabeled images during training or assume all unlabeled samples to belong in the same class.
- The number of image classes the OSLS is able to classify is greater than the ones used in the open-set classification literature [10, 7, 25].

## II. Feature Extractor

Deep Residual Networks have been proven to be a very effective in mapping images to a meaningful feature space, especially when trained from large datasets [32]. In this work we use ResNet18 and ResNet34 [16] to map the sample images to the vector space. Both architectures produce 512-long feature vectors which compared to deeper network feature products lead to a shorter algorithm running time. The different types of ResNets we used, although not very different, will be discussed in Section 3. The weights were trained using the ImageNet1k dataset [6] which involves a large-scale ontology of images. The development of the feature extractor itself is out of the scope of this paper and pre trained feature extractors available by PyTorch are used.

Before providing the training images feature vectors to the OSLS classifier we normalize them using the following equation:

---

[1] https://pytorch.org/vision/stable/models.html

$$F_{norm} = \frac{F - F_{min}}{F_{max} - F_{min}} \qquad (1)$$

Where $F$ is a 512-long vector feature map and $F_{max}, F_{min}$ are its respective maximum and minimum values in vector form. This type of basic normalization constrains the $F$ values from 0 to 1. We apply the normalization to prevent the Exponential Loss and its Derivative in Equation 9 and Equation 11, respectively, from gaining extremely high values. Additionally, we demonstrate in Section 6 that this type of normalization significantly increases our method's classification accuracy compared to the more popular softmax normalization. It can be argued that the use of softmax normalization yields poor solutions for open-set recognition as it tends to overfit on the training classes.

## III. Datasets

The proposed OSLS classification method is used on a variety of training and testing examples, each using different sample arrangements. To explore the capabilities of the proposed method in different settings, two different datasets are being used: the Caltech256 [11] and a custom Mixed dataset.

To explain with more clarity our method and results, we describe the way the Caltech256 dataset is split in two groups. Caltech256 is an open source dataset, it consists of 256 different image classes and has been recently used a lot as a benchmark for a variety of machine learning applications [2, 9].

Similar to our selection of ResNets, we use an open source and broadly used dataset in order to make our example and results as general and less task specific as possible. As we intend to produce work that is going to be used in the future for specific applications, to give a hint on how the method can be geared towards recognition in unique environments the eight infrared classes (Figure 3) which are available from the Military Sensing Information Analysis Center (SENSIAC) Automatic Target Recognition (ATR) database [34] are used in a number of our tests.

We train our classifier on both labeled and unlabeled pictures, therefore our main dataset consists of what we call Relevant and Irrelevant pictures. The relevant group is consisted of the first 50 classes of Caltech256 and the irrelevant group contains images from the next 50 classes as shown in Figure 2. To explore the dependency between our classifier visual recognition accuracy and the amount of unlabeled images, we created two more versions of the Caltech256 dataset with an expanded number of Irrelevant images, one has 100 classes of unlabeled images (+50 Irrelevant) and the other has 200 classes of unlabeled images (+150 Irrelevant), both with the same number of pictures per class, 40.

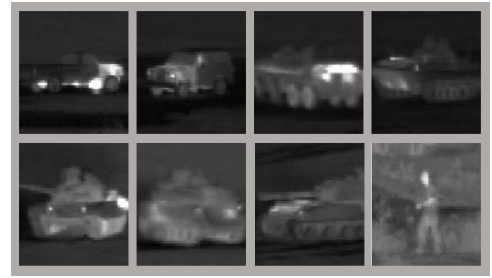

*Figure 3:* The manually created infrared (IR) dataset using video snapshots from the ATR database consists of eight classes, seven of them are civilian and combat vehicles and the last one is a human class.

Finally, in order to explore the behavior of our method on unique and very different environments from the ones present in the Caltech256 dataset, we created our own infrared (IR) combat vehicle image group by taking snapshots from the publicly available IR videos provided in the Military Sensing Information Analysis Center (SENSIAC) ATR database (examples displayed in Figure 3). The new data product is composed of the same amount of pictures with the one in Figure 2, with the exception that the first 8 relevant image classes are infrared instead of Caltech 256 pictures. Although only

Select Classes from Caltech 256 Dataset

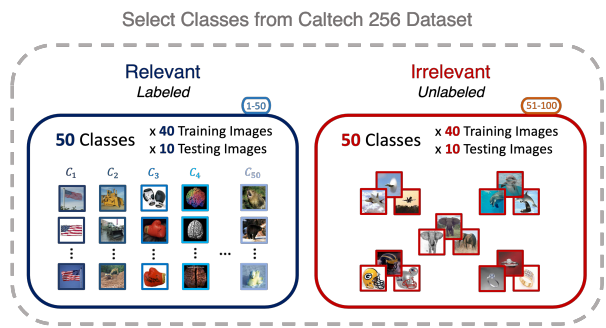

*Figure 2:* In blue we see the 50 Relevant image classes, and in red the Irrelevant. For training, every class, both labeled and unlabeled contains 40 images, so 2000 labeled and 2000 unlabeled. The evaluation is performed using 10 images for each class, different than the ones used during training.

Relevant

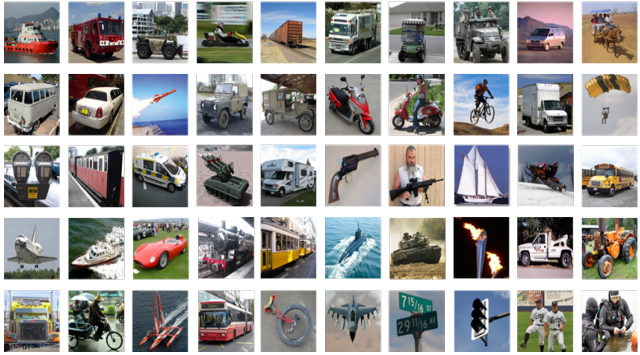

Irrelevant

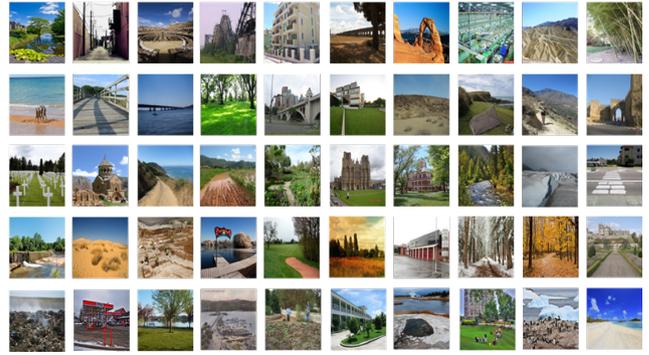

*Figure 4:* Example images from each one of the 100 classes that comprise the mixed dataset. The 50 relevant images include a variety of different vehicles, humans and weapons while the 50 irrelevant include buildings and outdoor scenery.

8 infrared classes are available in the ATR dataset, the term Infrared Dataset is used to indicate that 8 infrared classes are included within the 50 relevant classes. On the next chapter we discuss how the dataset images described above are treated during the classifier training process.

The second dataset we train and test our classification method on is the Mixed dataset. The relevant group is composed of 50 select classes from the ImageNet [6] dataset and includes pictures of vehicles, aircraft, humans and weapons. This group will serve as the target images that are expected to be recognised. On the other hand, the irrelevant group is composed of 50 classes from the MIT Places [35] dataset which includes a variety of outdoor scenery pictures such as buildings and natural environment. The choice of these two datasets is deliberate as in our application we are trying to recognize objects in a scene and push away scenes that have no relevant objects. Each class on the relevant part of the mixed dataset is comprised of 1,300 images, while each irrelevant class has available 13,000 pictures on average. The imbalance between relevant and irrelevant images is representative of the imbalance in the unlabeled data captured in the field which will contain many more irrelevant objects compared to targeted classes. From every class in both groups, 10 images are reserved for testing the accuracy of the various methods which are compared after the training of the classifier has been completed. When using the Mixed dataset in this work, a part of the irrelevant pictures will be reserved and used as unseen samples (Figure 9), images that have not been seen during training but have to be recognized by the classifier in the same way as the irrelevant.

Examples of the mixed dataset images are presented on Figure 4 while a complete list of the classes in alphabetical order is presented in the Appendix. The way the mixed dataset is utilised for training and testing the low-shot classifier is discussed in Section 5.

## IV. Low-Shot Classifier Training

The two integral parts of our classifier training process are the target matrix and the loss function. Our training goal is to tweak the initially randomized weight matrix in such a manner that when multiplying it with a testing feature map, it produces a score matrix whose largest value is the desired class element.

In machine learning, a fully connected layer performs the following calculation:

$$\hat{y} = \sigma(WF + \beta) \qquad (2)$$

Where W is weighting matrix of the classifier, F is the feature map matrix, V is the bias vector and f is the activation function. A Singular Value Decomposition (SVD) method solves our matrix equations [21, 19]. The pseudo-inverse method calculation results as:

$$\hat{y} = \sigma WF \qquad (3)$$

Here, $F$ is the feature maps,, is the weight matrix which we desire to train, and ^H is the target, the ideal outcome for the score matrix. Our MATLAB implementation handles the training one class at the time, therefore $F$ is a $N_{img} \times N_{feat}$ matrix and $W$, is a $N_{feat}$ long vector for each class, where $N_{img}$ is the number of training images in every epoch and $N_{feat}$ = 512 is the length of the feature vectors (constant). With no use of the bias vector, and the reversed order of $F$ and ,to account for the row-column switch, in SVD we calculate the $W$, matrix one vector (class) at a time therefore essentially solving for the least square solution of:

$$Ax = b \qquad (4)$$

When the exact solution does not exist, which means that $A$ is not a full-rank square matrix, we get approximate solutions as:

$$Ax = \hat{b} \qquad (5)$$

Therefore the approximation error is:

$$d = \hat{b} - b \tag{6}$$

and in a least-square approach the loss function is:

$$L = ||d|| = \sqrt{\sum_{i=1}^{n} d_i^2} = \sqrt{\sum_{i=1}^{n} (\hat{b}_i - b_i)^2} \tag{7}$$

substituting Equation 6 results:

$$L = \sqrt{\sum_{i=1}^{n} \left(\sum_{j=1}^{m} A_{ij} x_j - b_i\right)^2} \tag{8}$$

We introduce an exponential version of the least square solution in order to explore a new, faster converging loss function based on [22]. Our new squared-exponential loss function is:

$$L = \sum_{i=1}^{n} e^{d_i^2} = \sum_{i=1}^{n} e^{(\sum_{j=1}^{m} A_{ij} x_j - b_i)^2} \tag{9}$$

therefore the gradient can be proven analytically to be:

$$\frac{\partial L}{\partial x_j} = \sum_{i=1}^{n} 2e^{d_i^2} d_i A_{ij} \tag{10}$$

and the gradient vector is:

$$\frac{\partial L}{\partial x} = A^T (d. * e^{d^2}) \tag{11}$$

There are two main reasons for choosing this loss function. A squared-exponential function is easy to differentiate analytically and the differentiation is applied to the linear algebra form implemented in the MATLAB code. Note that the dot operator in Equation 11, i.e., $.*$, used with multiplication in MATLAB, creates element wise operations. Compared to other differentiable functions we tested, the square-exponential was the one to converge faster and in a steady way. A problem we encountered, which we solved by normalizing the feature maps as described in Equation 1, is that because of the nature of the function, for numbers greater than 1, the Loss would result in extremely high values.

The gradient matrix in Equation 11 is then multiplied by a learning rate($\eta$) and added to the weight matrix ($W$) repeating this sequence for every epoch. The steps taken towards training the classifier matrix are therefore all independent from machine learning libraries or functions. Although many different loss functions that get differentiated in a semi-analytic fashion are being used by machine learning libraries, we concentrated our efforts on not using any existing libraries to create a stand-alone method. Therefore the squared-exponential loss function is a good fit.

As in most machine learning applications, the update mechanism used towards convergence is some variation of a normal gradient descent equation. In our specific case we use:

$$W_{k+1} = W_k - \frac{1}{2}\eta \frac{\partial L}{\partial x} \tag{12}$$

Here, in every epoch $k$, $W$ gets updated by subtracting from it the product of the learning rate [ and the gradient matrix. We obtain our learning rate using an algorithm inspired by Iterative Shrinkage-Thresholding Algorithm (ISTA) [3]. We begin with calculating a pseudo-loss which is going to be compared with the actual loss to determine whether the learning rate needs to be decreased or kept as specified on the previous epoch. This iterative method progressively decreases the learning rate as we approach closer to the desired optimal point.

The last, and most unique part about our classification mAs in most machine learning applications, the update mechanism used towards convergence is some variation of a normal gradient descent equation. In our specific case we use: ethod is our target. As mentioned in the Section 1, the uniqueness of our approach relies on the fact that we make use of unlabeled images during the training of the weight matrix. This is done by extending a typical one-hot encoding [14] matrix to also include class score distributions as targets for the Irrelevant images. Labeled images have arrays of zeros and a unit value on the correct class element as targets.

Irrelevant pictures belong to none of the classes therefore the score for each class should be zero. By experimentation we concluded that the irrelevant target that works best should be a slight negative value, such as -0.2. This intuition matches some of the binary classification work that has been done on Support Vector Machines' (SVM) correlation filters, where 0 and 1 were not as separable as a negative value (-0.1, -1) and 1 [36, 5]. As an example, if a training dataset was consisted of six pictures, half of them labeled and half of them unlabeled, and the labeled ones were members of three different classes, our target matrix would look as follows:

$$\hat{y} = \begin{bmatrix} 1 & 0 & 0 \\ 0 & 1 & 0 \\ 0 & 0 & 1 \\ -0.2 & -0.2 & -0.2 \\ -0.2 & -0.2 & -0.2 \\ -0.2 & -0.2 & -0.2 \end{bmatrix} \tag{13}$$

We train the weight matrix in such a way that during evaluation the irrelevant images class score vector values are spread equally between the classes and acquire values as close to zero as possible. This helps the Irrelevant pictures to score less than the respective class threshold.

Along with using this target oriented training procedure, we also increase our recognition accuracy by calculating our threshold scores using the ROC method as explained in Section 5.

## V. Low-Shot Classifier and ROC Threshold Calculation

The multiplication between a feature vector and a weight matrix yields a class score vector. Neural Network classification theory uses the highest score (Top 1, Top 3 or Top 5 have been used too) to group the images into classes. We extend this criterion to make it applicable when unlabeled images are present by introducing a Threshold Score ($T_S$) value for each class.

The $T_S$ value serves as a binary discriminating test in order to group pictures in the Relevant and Irrelevant bins. As mentioned above, we don't only need to divide pictures in the two groups, we also want the relevant group pictures to be normally classified in their respective class.

It is important to note that the $T_S$ is calculated during the training of the OSLS. We need the $T_S$ to be pre-calculated before we start evaluating our testing dataset. Once the training has been completed and the $T_S$ is known, the classification process runs as follows: a) The testing image runs through the classifier and scores, which denote the likelihood of the image belonging to each class, are calculated. b) The image is assigned to the class with highest score. c) The score of the assigned image is compared to the $T_S$ of the class where it was assigned. If it is higher then it is considered as a member of the class. If lower, it is determined to be an irrelevant image.

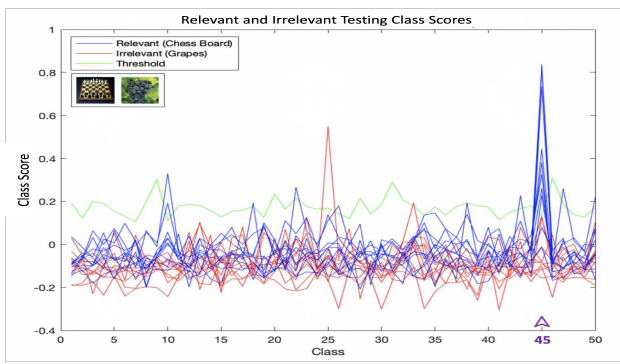

*Figure 5:* In blue and red we see the 50 different class score values of the 10 different Chess Board and Grapes images respectively. In Green is the ROC threshold calculated for each one of the 50 classes.

Within Figure 5 we present an example where class score values for ten testing images from two classes are plotted. During the training of the classifier matrix we treated the Chess Board pictures as labeled (Relevant), with label 45 attributed to them, and the Grapes pictures as unlabeled (Irrelevant). On the graph we can see that during testing, most of the Chess Board pictures have high class score values on the correct class (45). The correct scores are also above the $T_S$ of this class, therefore they will be classified correctly as members of the chess board class.

On the other hand, the Grapes pictures are treated as Irrelevant during training resulting in lower score values compared to the $T_S$ (green line) of all relevant values. Our target matrix along with our classification method achieves to push the Irrelevant score values lower than the Relevant (red lines below the green line), while keeping the score values for the relevant class above the corresponding $T_S$ (blue lines above green in class 45). Intuitively, we can set the $T_S$ to be the lowest relevant value encountered during the training (Normal Threshold). If a picture is not classified higher than the worst correctly classified training picture, then it should be Irrelevant. As seen in Figure 6, although this discriminatory rule will give us the best possible Relevant accuracy, it will strongly discriminate against Irrelevant pictures.

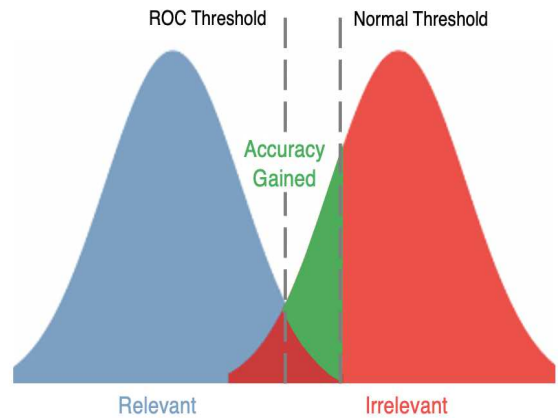

*Figure 6:* In blue we see the normal distribution of Relevant scores within class X, while in red we see the distribution of the Irrelevant pictures that got classified as class X. Using a normal threshold would classify all Relevant pictures as True Positive, but would hurt the True Negative and total accuracy by a value represented by the green area.

To maximize the combined accuracy we use the ROC curve [27, 13, 12] to chose our threshold values. The same Way we would do with a Normal Threshold, the ROC Threshold is going to be calculated right after training and before testing, using explicitly the training data.

To demonstrate the need of using the ROC $T_S$, we graph the ROC curves of five, unique compared to each other, classes of pictures that we trained our classifier on. All five classes were part of the labeled dataset (relevant classes). We can see in Figure 7 that every different class of pictures has a 6 different response to the ROC implementation. The different Areas Under the Curves (AUC) represent how well our ROC method is able to classify the data, but also underlines the need of such an implementation.

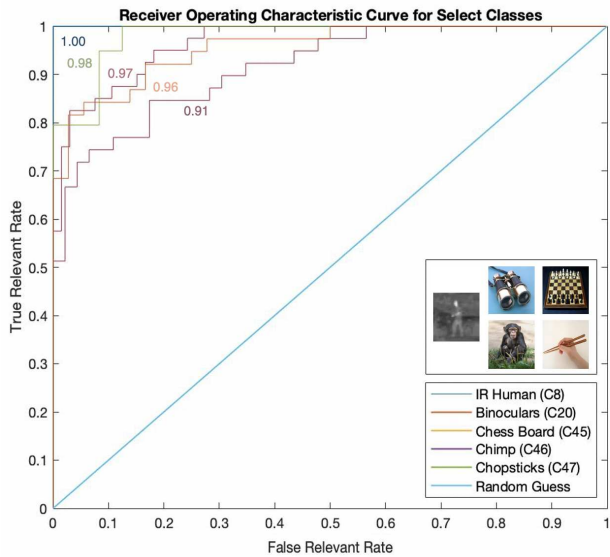

*Figure 7:* True Relevant Rate is the amount of relevant pictures that our classifier recognized them as such, over the total number of correctly classified pictures. False Relevant is the ratio of pictures that were irrelevant but were predicted as relevant, over the total number of incorrectly classified pictures.

In our analysis we focus on the classification of Relevant pictures, which we assume to be the Positive statistical case, therefore we use the terms True Relevant Rate (TRR) and False Relevant Rate (FRR). TRR and FRR are no different than the True Positive Rate (or Sensitivity) and False Positive Rate (or Fall-Out) respectively, used in statistical analysis.

Therefore we define:

$$TRR = \frac{TP}{TP+FN} \quad \text{and} \quad FRR = \frac{FP}{FP+TN} \quad (14)$$

In Figure 7 it is obvious that the IR Human class is so unique that the classifier does not have any trouble distinguishing it from the rest of the dataset, therefore as seen in the figure above its AUC equals to 1 and the ROC does not have much effect on its cumulative accuracy. Different classes though present different levels of difficulty for our classifier. The Chimp class as seen above, has an AUC of 0.91, which means that the ROC can significantly improve its cumulative accuracy if a $T_S$ is picked wisely.

The Normal Threshold would pick the point on the graph where the False Relevant Rate is minimum for a True Relevant Rate of 1, hence, for the Chimp graph, the [0.57, 1.00°] point. Using our ROC algorithm we can pick any other point on the graph, such as the [0.20, 0.84°] point which is the one further away from the blue line that represents a random guess. By doing this, although we slightly decreased our TRR, we get a great increase in FRR, which results to a significantly higher cumulative accuracy. This accuracy increase is going to be clearly presented and discussed in the next Section.

Tables 1, 2 and 3 demonstrate the usefulness of the ROC method in classifying Relevant images and rejecting an image if it is Irrelevant. In order to highlight the ROC capabilities, we compare our results to the two baseline methods, the "No Irrelevant" and the "+1 Class".

The first baseline result (No Irrelevant), was produced by training the classifier only on labeled images of the relevant classes. This is the case where although we have unlabeled images for the irrelevant classes, we do not use them, expecting the labeled images to have enough meaningful features to accommodate recognizing the irrelevant ones. To evaluate this method, we use our Normal Threshold Score Criterion discussed above where we set the lowest correct relevant training score as the threshold for each class. During testing, if the image's highest class score is larger than the respective threshold then its classified as Relevant, if not as irrelevant.

The second baseline result, which we call "+1 Class", was generated by training the classifier to recognize the relevant classes along with one extra class which encapsulates all irrelevant images. During the training of the classifier, all unlabeled images of the Irrelevant classes were assigned to an extra class. The evaluation is being done by simply comparing the highest scoring index of every image with the correct target.

Table 1 shows how the baseline methods scored for both relevant and irrelevant images compared to the Low-Shot Classifier, with and without applying the ROC optimization for the Top-1 selections.

*Table 1:* Low-Shot Classifier Compared to Baseline Examples for the Top-1 Selections

| Classifier | Normal Dataset | | Infrared Dataset | |
|---|---|---|---|---|
| | R | I | R | I |
| Low-Shot Classifier | 70.8 % | 74.8 % | 78.2 % | 75.4 % |
| Low-Shot Classifier w/ ROC | 64.8 % | 87.8 % | 71.8 % | 89.8 % |
| +1 Class | 49.2 % | 91.4 % | 56.2 % | 92.2 % |
| No Irrelevant | 72.4 % | 47.8 % | 78.6 % | 52.4 % |

As "Normal" we describe the dataset consisted of 50 relevant and 50 irrelevant Caltech256 classes and as "Infrared" the dataset where we have substituted 8 of the relevant classes with IR ones. Both datasets are described in Section 3. The Low-Shot Classifier results are obtained by running the algorithm described up until Section 6 and the Low-Shot Classifier with ROC by adding the ROC extension. "R" and "I" are the relevant and irrelevant classification accuracy re-spectively. The numbers shown in the tables are the Top-1 percentages of images that got classified correctly during evaluation.

It can be observed that the baseline methods are unable to classify decently both groups of images. The +1 Class method seems to over-train the classifier on recognising the unlabeled images failing to put the labeled ones in the correct classes. This happens most likely due to the unbalanced training data, as the 51st class has as many images as the rest 50 together. On the other hand, by using only labeled images, we train the classifier to specifically recognise the labeled group, failing to filter out the unlabeled images "noise".

In the first row of Table 1, the results of our classifier without the ROC extension show that our loss function combined with our unique target matrix and the threshold score criterion can recognize equally well both labeled and unlabeled images. It is notable that for the relevant group our method loses a small amount of accuracy compared to the label-specific baseline method but does substantially better in identifying irrelevant images.

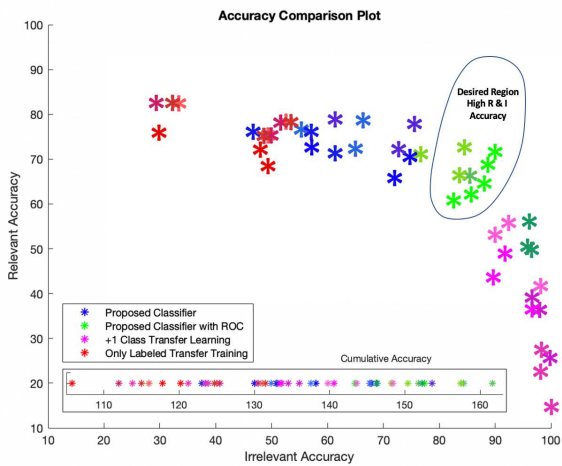

*Figure 8:* The legend follows the color coding of the four different methods in Table 1. For each of the four methods the graph has four different sub color groups with three data points each. The four different sub groups represent the different datasets discussed in Section 2 and the data points are the three different pre-trained feature extractors used.

The ROC method greatly increases the unlabeled images recognition, to the modest expense of the labeled images. The table shows the importance of using the ROC to greatly increase the cumulative accuracy. Our ROC classifier increases by 12% the cumulative recognition scores compared to the +1 Class method and by 25.4% compared to the label exclusive transfer learning method.

The described results are also depicted in the Accuracy Comparison Graph in Figure 8. For every method discussed we use three different ResNet10 feature extractors (BatchSGM, SGM, L2) in order to show the consistency of our classifier results.

With a few exceptions, no matter the feature extractor or the nature of our dataset (including infrared, including more unlabeled images), our proposed method (green data points) not only provides a higher cumulative accuracy but also eliminates the bias between labeled and unlabeled images by classifying both equally well when compared to the baseline approaches. In the graph we introduced the results of our extended datasets which consists of more unlabeled images.

Table 2 offers a closer look to the comparison of the two extended datasets.

*Table 2:* Extended Datasets Comparisons

|  | + 50 Irrelevant | | + 150 Irrelevant | |
| --- | --- | --- | --- | --- |
| Classifier | R | I | R | I |
| Low-Shot Classifier | 79.0 % | 66.3 % | 76.4 % | 56.7 % |
| Low-Shot Classifier w\ ROC | 73.0 % | 84.4 % | 59.0 % | 93.0 % |
| +1 Class | 36.8 % | 97.7 % | 22.0 % | 99.2 % |
| No Irrelevant | 78.6 % | 51.9 % | 78.6 % | 52.9 % |

We follow the same notation in Table 2 as used in Table 1, with the only difference being that the "+50" and "+150" Irrelevant datasets are the two expanded datasets noted in Section 3. Although it is clear in both in Table 2 and the plot in Figure 8, that our method still scores better in a cumulative perspective, we can also observe that biases against the relevant (in Low-Shot Classifier with ROC algorithm) or irrelevant (in Low-Shot Classifier algorithm) group begin to occur when increasing the amount of irrelevant images.

We see that for the Low-Shot Classifier the more we increase the irrelevant to relevant ratio the worse we score on the irrelevant part. This might seem counter intuitive as we would expect that the more unlabeled images we see during training, the better we would be able to recognise them. In reality, we introduce many more feature elements on the irrelevant part, which leads to consequently eliminating their uniqueness.

When introducing the proposed ROC approach on the second row (Table 2) we do not observe the introduction of bias because the ROC threshold has been adjusted in such a way that it is non discriminating against any group (OptimalROC). On Table 3 we present the results of our adjusted ROC Classifier being used on the +150 Irrelevant dataset. The same behavior is observed when we test the rest datasets.

The "Optimal ROC" and "No Irrelevant" rows correspond to Table 2 second and fourth data rows. Putting a constraint on how much we are willing to shift the )( to limit the loss in relevant, affects negatively the irrelevant. We desire to find a percentage which during testing gives us a decent cumulative accuracy without big losses on the Relevant part. This could be imagined as turning a knob to tune our ROC implementation. This can be specific in every application, therefore an open ended approach is adopted.

*Table 3:* ROC Adjustment for the +150 Irrelevant dataset

|  | R | I |  |
|---|---|---|---|
| No Irrelevant | 78.6 % | 52.9 % | Increase |
| Optimal ROC | 59.0 % | 93.0 % | + 20.5 % |
| 80% Constraint | 61.4 % | 90.2 % | + 20.1 % |
| 90% Constraint | 68.2 % | 82.0 % | + 18.7 % |
| 92.5% Constraint | 72.0 % | 77.0 % | + 17.5 % |

A 100% constraint would be the Low-Shot Classifier without ROC, as we set our Threshold Scores to be the lowest correctly classified irrelevant picture in every class. On the table presented, a 90% Constraint means that we ask our ROC algorithm to keep our thresholds to a value that will not hurt our correct relevant guesses more than 10% during the calculation of the $T_S$. Therefore these constraints are applied when using the training images and they differ from the percentages encountered in the testing (Table 3). As we can see, for the specific case we can compromise with an 18.7% total increase instead of the the 20.5% of the optimal case, in order to get a more equal recognition accuracy.

## VI. OSLS Classifier Results

The Low-Shot Classifier is able to recognise images from the relevant classes and also identify irrelevant images from the classes it has seen during training. Ideally, during operation we desire to recognise objects that are not seen at all during training, which is the main objective of openset recognition. To achieve this we extend the capabilities of the Low-Shot Classifier described in Section 3 to recognizing unseen images resulting the OSLS Classifier. The unseen samples are the sub-group of the irrelevant classes that do not get involved in training but it is still expected that the OSLS Classifier recognises them as irrelevant. This is accomplished by randomizing the selection of the irrelevant samples in every training epoch of the OSLS classifier. More specifically, during the training of the Low-Shot Classifier, there are 2 number of classes each of which contain= number of training images for both relevant and irrelevant datasets (only the images for the relevant dataset are labeled). This set of $n \times c$ images is the same in each epoch. The difference in training the Open-Set Low-Shot Classifier seen in Figure 9 is that the irrelevant images are different in each epoch, and selected randomly from the pool of unlabeled irrelevant images while still keeping the total number of irrelevant training images in each epoch the same with the relevant part ($n \times c$). By introducing this imbalance and by not repeating the same irrelevant samples in each epoch, our classifier is able to generalize better on the irrelevant part, yielding better classification accuracy for the irrelevant and unseen testing samples. In all results presented in this paper the testing images are always different than the images used during training.

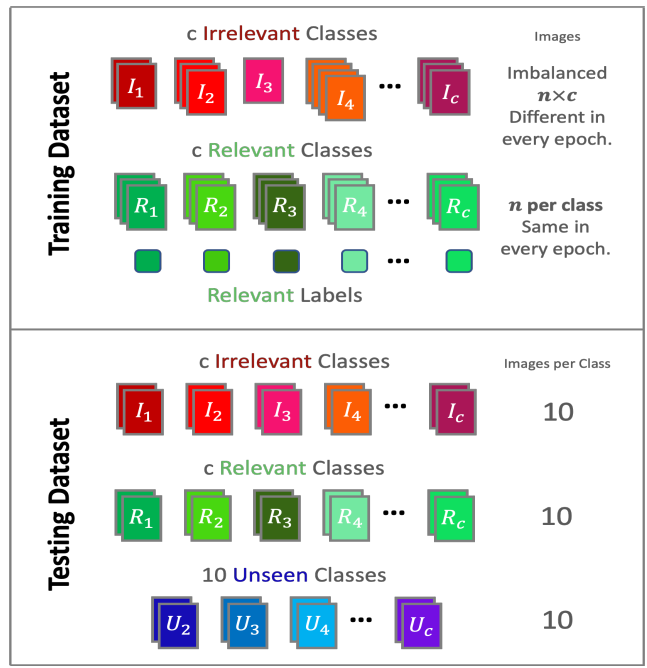

*Figure 9:* The two differences of the OSLS method compared to the Low-Shot Classifier: We extend our testing dataset to include 10 classes of unseen images (each containing 10 testing samples) while also extending the irrelevant part of the training dataset by introducing randomness and imbalance between epochs and classes respectively. During training, a different selection of irrelevant samples and the same selection of relevant samples is used in every epoch.

A comparison between the traditional (Low-Shot Classifier) and the randomized irrelevant training of the Low-Shot Open-Set Classifier is presented in Figure 10. In this examination, the Low-Shot Classifier training uses the same relevant and irrelevant pictures and classes (40) in every training epoch, whereas the OSLS uses the same set of relevant classes (40) and pictures but samples randomly a different group of irrelevant training images in every epoch. For instance, in the 40 images per class case, the Low-Shot Classifier is trained on the same 40 relevant and irrelevant classes which all include the same 40 pictures for each class. For the OSLS Classifier case, although the 1,600 relevant images (from 40 different classes) are kept the same throughout training, the 1,600 irrelevant images in each epoch are picked randomly from a pool of 20,000 samples (500 samples for each one of the 40 classes), introducing not only randomness but also imbalance between class samples.

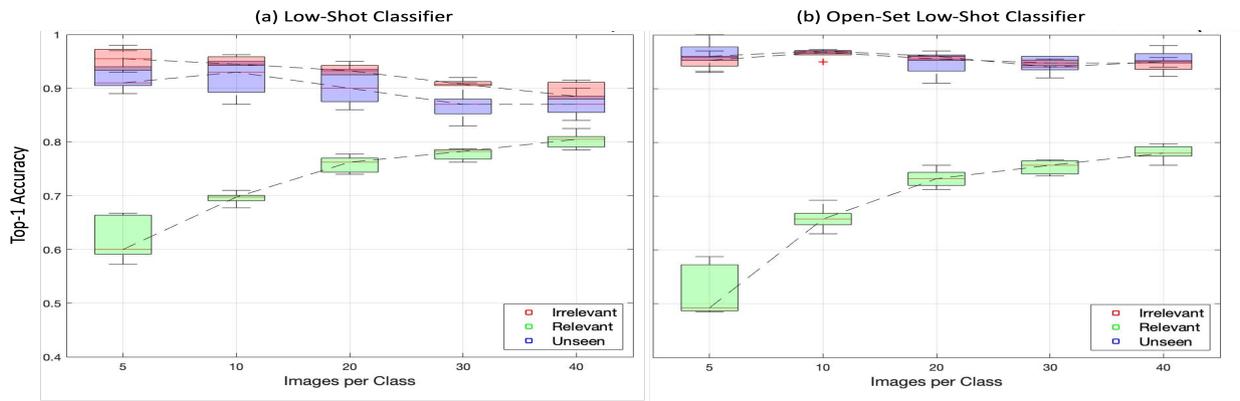

*Figure 10:* The accuracy box plots for the three different testing groups are presented: in red the Irrelevant, in blue the Unseen and in green the Relevant results. For every Images per Class case, ten different random tests are performed in order to quantify the uncertainty of each case study. The box plot sides represent the median of the lower and upper half of the different results set respectively. The lines extending from the boxes (whiskers) indicate the variability outside the upper and lower quartiles while the red line within the boxes represents the median of the entire spread. Lastly, the red crosses represent the accuracy of the outlier runs and the dashed line connects the median accuracy values of all the different cases.

Figure 10 demonstrates that by introducing randomness and class imbalance during training, for every Images per Class case there is a slight decrease in the relevant accuracy, but a substantial increase in the testing performance of both irrelevant and unseen. All the results below use the ROC threshold that produces the highest combined relevant and irrelevant score.

All the results presented in the box plots of this text are for an OSLS classifier that is trained on 40 relevant and 40 irrelevant classes, both of which have the amount of relevant images per class specified in the x-axis. Similar works in the open-set literature [10, 7, 25] are using a lower number of classes during training and testing (10 to 95 classes compared to the total number of classes used in this paper ranging between 90 and 250). To show how the OSLS Classifier performs in tests where the same order of classes are used, we vary the number of relevant and irrelevant classes used during training. Although it is of interest to recognize samples of as many classes as possible (a maximum of 40 as presented in Figure 10), by observing Table 4 it is evident that the OSLS Classifier achieves very high Top-1 accuracy scores in situations where the relevant and irrelevant classes we are trying to detect are limited.

We take as an example the case (in bold) where we train the classifier on 10 Relevant and 10 Irrelevant classes each one 10 of which includes 40 training samples- a total of 400 labeled and 400 unlabeled images. By testing using 10 samples per class from 10 relevant, 10 irrelevant and 10 unseen classes the classifier achieves Top-1 accuracy scores of 0.89 ± 0.04, 0.98 ± 0.02 and 0.96 ± 0.02 respectively, with a very low variance between the random runs (f ≤ 0.04).

*Table 4:* The complete set of results for the OSLS method for a variable number of classes and images per class. Horizontally are presented the results for a variable number of pictures per class (P). Vertically are presented the results for a variable number of classes (C) for each one of the three testing sample categories. For each different example, the mean and standard deviation of 10 different random tests is presented for the Top-1 accuracy.

| P<br>C | 5 | 10 | 20 | 30 | 40 |
|---|---|---|---|---|---|
| *Relevant* | | | | | |
| 5 | 0.56 ± 0.12 | 0.61 ± 0.07 | 0.75 ± 0.09 | 0.78 ± 0.03 | 0.83 ± 0.08 |
| 10 | 0.53 ± 0.07 | 0.66 ± 0.07 | 0.83 ± 0.03 | 0.84 ± 0.03 | **0.89 ± 0.04** |
| 20 | 0.53 ± 0.07 | 0.66 ± 0.02 | 0.78 ± 0.02 | 0.83 ± 0.01 | 0.84 ± 0.03 |
| 30 | 0.51 ± 0.05 | 0.66 ± 0.03 | 0.74 ± 0.02 | 0.77 ± 0.02 | 0.78 ± 0.01 |
| 40 | 0.51 ± 0.05 | 0.66 ± 0.02 | 0.73 ± 0.02 | 0.75 ± 0.01 | 0.76 ± 0.02 |
| *Irrelevant* | | | | | |
| 5 | 0.95 ± 0.08 | 0.97 ± 0.03 | 0.98 ± 0.02 | 0.98 ± 0.02 | 0.99 ± 0.01 |
| 10 | 0.93 ± 0.04 | 0.97 ± 0.04 | 0.97 ± 0.01 | 0.98 ± 0.01 | **0.98 ± 0.02** |
| 20 | 0.95 ± 0.04 | 0.98 ± 0.01 | 0.97 ± 0.01 | 0.96 ± 0.01 | 0.97 ± 0.02 |
| 30 | 0.97 ± 0.02 | 0.98 ± 0.01 | 0.97 ± 0.01 | 0.96 ± 0.01 | 0.96 ± 0.01 |
| 40 | 0.96 ± 0.02 | 0.95 ± 0.01 | 0.95 ± 0.01 | 0.95 ± 0.01 | 0.94 ± 0.01 |
| *Unseen* | | | | | |
| 5 | 0.93 ± 0.06 | 0.96 ± 0.02 | 0.97 ± 0.03 | 0.98 ± 0.02 | 0.98 ± 0.01 |
| 10 | 0.95 ± 0.02 | 0.95 ± 0.04 | 0.97 ± 0.02 | 0.96 ± 0.02 | **0.96 ± 0.02** |
| 20 | 0.95 ± 0.05 | 0.98 ± 0.01 | 0.97 ± 0.01 | 0.95 ± 0.03 | 0.96 ± 0.01 |
| 30 | 0.96 ± 0.02 | 0.97 ± 0.02 | 0.97 ± 0.01 | 0.95 ± 0.02 | 0.93 ± 0.01 |
| 40 | 0.96 ± 0.02 | 0.96 ± 0.02 | 0.95 ± 0.02 | 0.95 ± 0.02 | 0.94 ± 0.01 |

The OSLS classifier is meant to be used as the final layer of any CNN that is expected to recognise samples that belong to the training classes while identifying as irrelevant images that are not relevant regardless if they originate from seen or unseen during training datasets. In order to demonstrate the versatility of the proposed method we attach the classifier to deeper feature extractors. Throughout the paper the feature extractor used to test any classifier was a pre-trained ResNet18 provided by PyTorch1. In Figure 11 results for a classifier similar to the one in Figure 10 are presented with the only difference being that the feature vectors are produced using the deeper ResNet34. Improvements in accuracy ranging from _ 0.11 to _ 0.02 (for the 5 and 40 Images per Class cases respectively), compared to those of Figure 10b, can be observed for the relevant testing samples while virtually no improvement is observed for the irrelevant and the unseen samples. Similar results are expected if the OSLS Classifier is used as a head for deeper networks which produce feature vectors of higher quality. The improvements can be attributed to the fact that a deeper network has the ability to produce better quality feature representations.

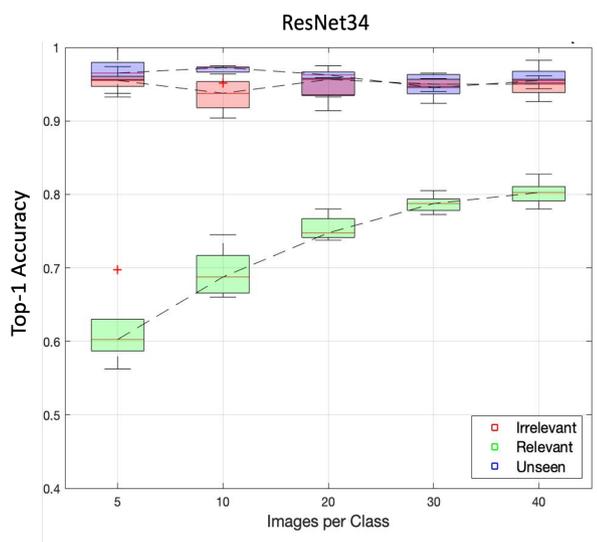

*Figure 11:* Box plots for the LSOS classifier presented in Figure 10b if the deeper ResNet34 is used to reduce the image samples to feature vectors.

The image feature representations used in this study are obtained raw, before any normalization is applied to them. As mentioned in Section 2, we use Equation 1 to normalized the input feature vectors. Figure 12 exhibits a decrease in accuracy if the features are normalized using the popular Softmax normalization commonly used in classification layers.

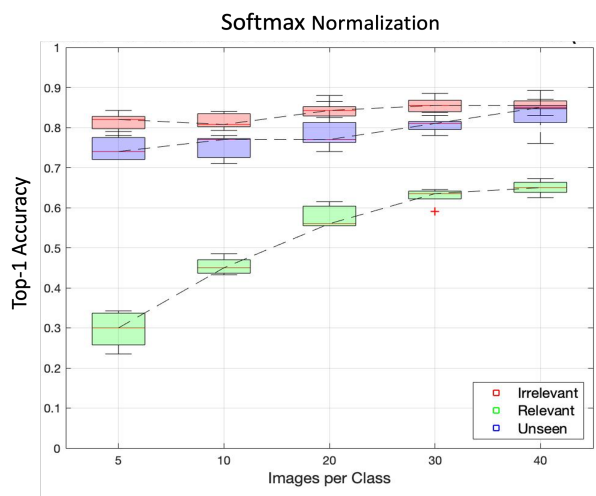

*Figure 12:* Box plots for the OSLS Classifier presented in Figure 10b if Softmax was used to normalize the training and testing samples.

More specifically, the OSLS results in Figure 10b show an improvement compared to Figure 12 that ranges from $\leq 0.19$ to $\geq 0.15$ for the relevant, $\leq 0.17$ to$\geq 0.08$ for the irrelevant and $\leq 0.26$ to $\geq 0.08$ for the unseen testing samples (for the 5 and 40 Images per Class cases respectively).

Finally, in order to demonstrate the value of the OSLS classifier, we compare it to the two baseline examples mentioned in Table 1. The first alternative method (Figure 13a) for classifying relevant samples along with rejecting irrelevant and unseen images is to group all the later in one class during training by assigning the extra class label to them ("+1 Class"). The second method (Figure 13b) the OSLS Classifier is compared to is a normal classification layer which is trained only on relevant images but is expected to recognize irrelevant and unseen images too ("No Irrelevant").

By comparing Figure 10b to Figure 13a, for a low number of samples per class, the "+1 Class" method performs equally well or in cases even better in all three categories compared to OSLS, with relevant accuracy scores ranging from $\geq 0.6$ to $\leq 0.7$ for the 5, 10 and 20 Images per Class cases while unseen and irrelevant recognition reaching accuracies $\leq 0.96$. When enough data samples per class are available though, the OSLS method improves the relevant accuracy by 0.05 and 0.1 for the 30 and 40 images per class cases respectively. The improvement in relevant image classification that the OSLS classification (Figure 10b) achieves is significant compared to the minor ($\approx 0.01$) decrease in relevant and unseen accuracy scores.

The benefits of introducing irrelevant images during trainingare evident when comparing the results of Figure 13b (No

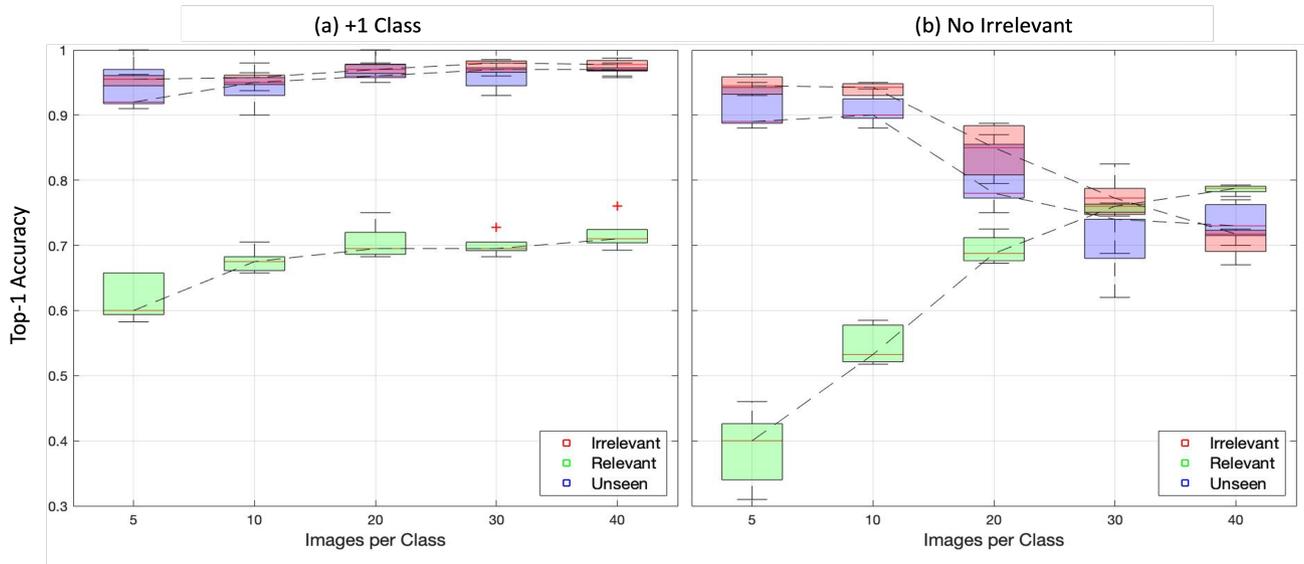

*Figure 13:* **(a)** The +1 Class method groups all the irrelevant samples in one class during training and expects the irrelevant and unseen testing samples to be classified like they belong to the extra class. **(b)** The No Irrelevant method is a normal classification layer which is trained only on relevant images although expected to recognise irrelevant and unseen images too.

Irrelevant) to the OSLS Classifier results in Figure 10b. If no irrelevant images are available during training and the number of relevant training images per class is small (5 and 10), a normal classification layer tends to over-fit on the later. Due to this over-fitting, the ROC Threshold rejects most of the samples during testing resulting to very high (≥ 0.9) irrelevant and unseen and very low (≤ 0.53) relevant accuracy scores. When there are more training images per class, a significant increase in relevant accuracy can be observed which is followed by a decrease in irrelevant and unseen accuracy. More specifically, assuming similar specifications (normalization, loss function etc.), if a single layer classifier is trained on 40 classes, each one including 40 images, the mean relevant, irrelevant and unseen accuracies during testing are 0.78, 0.72 and 0.73 respectively. If an equal number of unlabeled images are used during training, in the manner specified by the OSLS method, the mean relevant accuracy decreases by 0.02 while the irrelevant and unseen accuracy scores increase by 0.22 and 0.21 respectively. The trade-off between a very small decrease in relevant accuracy and a ten times larger increase in both irrelevant and unseen classification performance is the best demonstration of how the OSLS Classifier can be utilized in real-life applications.

The proposed OSLS Classifier using the ROC Threshold Score criterion not only makes the resulting model more flexible and easy to customize depending on the needs of the datasets, but also makes the method flexible for any application. This is a specifically interesting feature of our work, as we can use the classifier as an extension to any image recognition algorithm which desires to filter out irrelevant and unseen images without the expense of labeling.

## VII. Conclusion

In military reconnaissance applications a capability is needed where objects of interest -such as adversary targets- are reliably distinguished from objects of no relevance. Although a modest amount of labeled examples for the targets might be available to use during the training of the classifier, labels for the irrelevant objects might be scarce or even not possible to obtain.

To tackle this problem in this work we present an Open-Set Low-Shot Classifier which is trained using a modest number of labeled images from the relevant classes and unlabeled irrelevant images. A partially labeled target matrix is used for developing an analytically differentiated loss function for training the classifier. At each training epoch a random selection of the irrelevant images used in the training is introduced. During the training an ROC approach is used for determining a threshold score value for each relevant class. The latter is used for providing a balanced performance between classifying relevant samples and identifying irrelevant images.

During testing, this information is used for determining when a candidate image is either relevant, irrelevant or even unseen during training. The OSLS Classifier performs better compared to baseline classifying approaches, is able to handle the classification of many more classes compared to similar open-set approaches in the visual recognition literature and is able to demonstrate sufficient balance with high accuracy in classifying relevant images and identifying irrelevant images.

The code we based our experimentation and results on is available at: https://github.com/skasapis/ROCUnlabeledClassification

## Acknowledgements

We acknowledge the technical and financial support of the Automotive Research Center (ARC) in the University of Michigan in accordance with the Cooperative Agreement W56HZV-19-2-0001 of the U.S. Army CCDC Ground Vehicle Systems Center (GVSC) in Warren, MI.